\documentclass{article}
\usepackage{ijcai24}
\usepackage{makecell}

\usepackage{times}
\usepackage{soul}
\usepackage{url}
\usepackage[hidelinks]{hyperref}
\usepackage[utf8]{inputenc}
\usepackage[small]{caption}
\usepackage{graphicx}
\usepackage{amsmath}
\usepackage{amsthm}
\usepackage{booktabs}
\usepackage{algorithm}
\usepackage{algorithmic}
\usepackage[switch]{lineno}
\usepackage{stfloats}
\usepackage{multirow}

\title{Trustworthy Self-Attention: Enabling the Network to Focus Only on the Most Relevant References}

\author{
Yu Jing$^1$
\and
Tan Yujuan$^2$\and
Ren Ao$^3$\and
Liu Duo$^4$\\
\affiliations
$^{1,2,3,4}$Chongqing University\\
\emails
\{yu\_jing, tanyujuan, ren.ao, liuduo\}@cqu.edu.cn
}

\begin{document}

\maketitle
\begin{abstract}
    \renewcommand\floatpagefraction{.99}
    \renewcommand\topfraction{.99}
    \renewcommand\bottomfraction{.99}
    \begin{figure*}[b]
        \centering
        \includegraphics[width=16cm,]{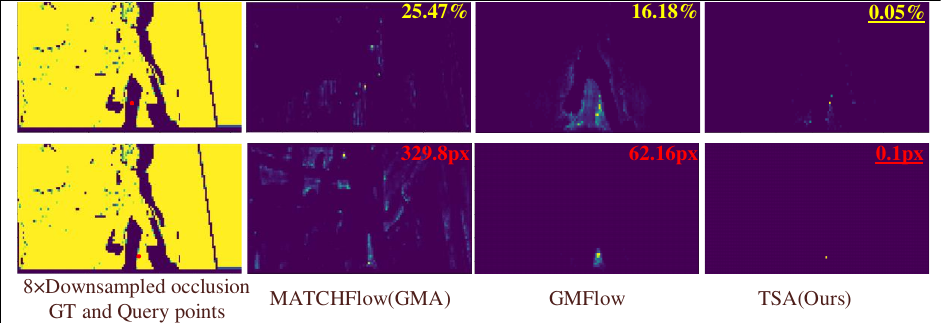} 
        \caption{The columns from left to right are the occluded ground truths of 1/8 downsampling, the attention maps of MATCHFlow, GMFlow, and TSA corresponding to the query points. The first row of yellow numbers represents the sum of the attention of all occluded points in the attention map corresponding to the occlusion ground truth (GT) with a red-marked occluded query point. It can be seen that our TSA can completely ignore occluded points \textbf{(Trustworthy)}. The second row of red numbers represents the Euclidean distance weighted sum between the point and its all reference points based on the attention map, corresponding to the red-marked non-occluded query points on the occlusion GT. It can be seen that our TSA can have much more focused attention, where non-occluded points only focus on themselves \textbf{(The most relevant)} (the attention of occluded points is also more concentrated \textbf{(Strongly relevant)}). The network used in this experiment was trained on Chair and Flythings datasets, and tested on the Sintel dataset.}
        \label{fig1}
    \end{figure*}
    The prediction of optical flow for occluded points is still a difficult problem that has not yet been solved. Recent methods use self-attention to find relevant non-occluded points as references for estimating the optical flow of occluded points based on the assumption of self-similarity. However, they rely on visual features of a single image and weak constraints, which are not sufficient to constrain the trained network to focus on erroneous and weakly relevant reference points. We make full use of online occlusion recognition information to construct occlusion extended visual features and two strong constraints, allowing the network to learn to focus only on the most relevant references without requiring occlusion ground truth to participate in the training of the network. Our method adds very few network parameters to the original framework, making it very lightweight. Extensive experiments show that our model has the greatest cross-dataset generalization. Our method achieves much greater error reduction, 18.6\%, 16.2\%, and 20.1\% for all points, non-occluded points, and occluded points respectively from the state-of-the-art GMA-base method, MATCHFlow(GMA), on Sintel Albedo pass. Furthermore, our model achieves state-of-the-art performance on the Sintel bench-marks, ranking \#1 among all published methods on Sintel clean pass. The code will be open-source.
\end{abstract}

\section{Introduction}

\renewcommand\floatpagefraction{.999999}
\renewcommand\topfraction{.9999}
\renewcommand\bottomfraction{.99999}
\renewcommand{\dblfloatpagefraction}{.9}
\begin{figure*}[htbp]
    \centering
    \includegraphics[width=17.5cm,height=7.9cm]{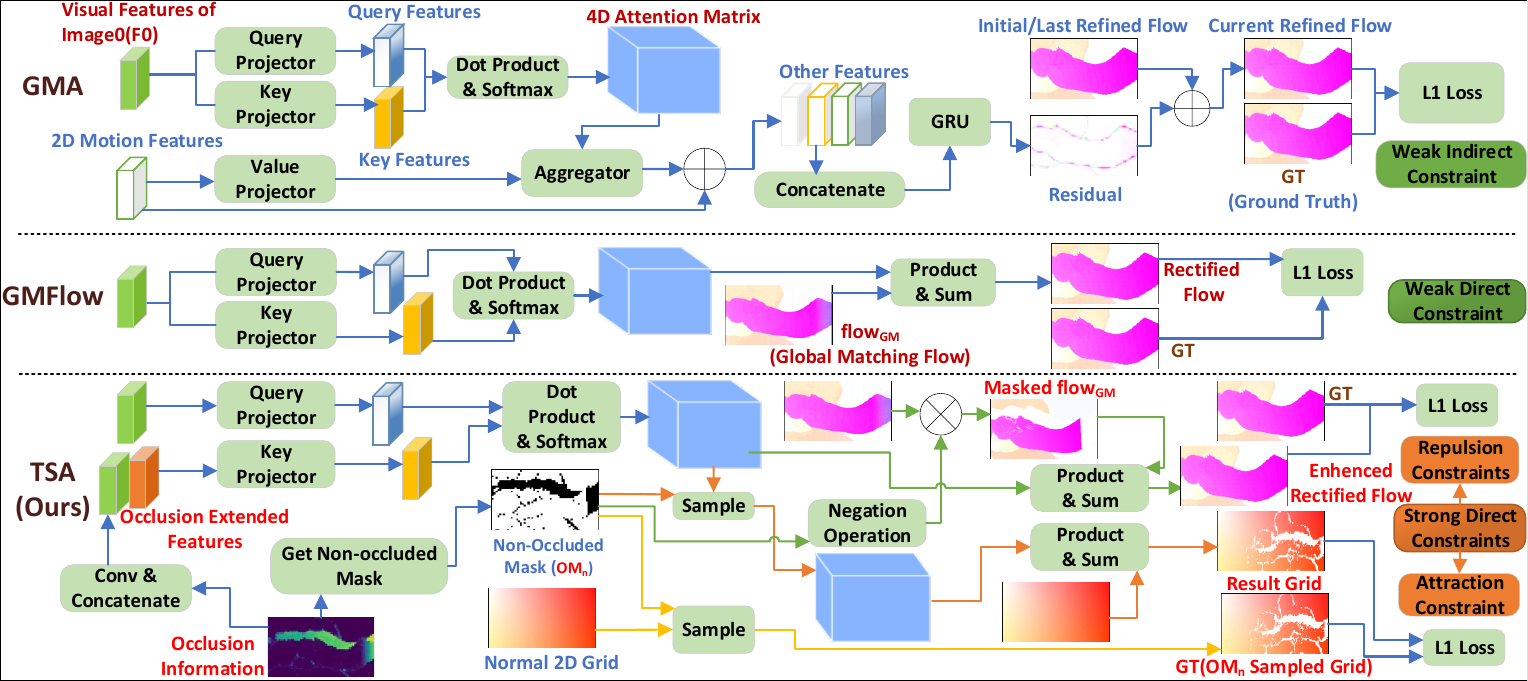}
    \caption{Comparison of three methods for training and reasoning of self-attention. Firstly, both GMA and GMFlow utilize self-attention solely on visual features of image0  to obtain a 4D attention matrix. However, due to the insufficient occlusion information carried by the image0 features, self-attention cannot effectively identify and exclude the focus on occluded points. This results in erroneous reference information obtained from this attention mechanism \textbf{(Untrustworthy)}. Therefore, our method addresses this issue by incorporating occlusion information into the self-attention to create occlusion extended features \textbf{(Trustworthy)}. Secondly, GMA utilizes weak indirect constraints to train, while GMFlow employs weak direct constraints. This results in better attention performance achieved by GMFlow. However, there is still the issue of attention scattering, where non-occluded points do not solely attend to themselves, leading to inaccurate estimation of optical flow on the surfaces of common objects that have different depths of field or are non-rigid \textbf{(Weakly relevant)}. Our method solves the problem of attention scattering by introducing two strong direct constraints, ensuring that non-occluded points only attend to themselves \textbf{(The most relevant)}. The performance of these methods are illustrated in Figure 1.}
    \label{fig2}
\end{figure*}
The optical flow of occluded points provides key basic information for other upstream tasks, such as 3D reconstruction from videoo \cite{yang2021viser}, video frame interpolation \cite{reda2022film}, object tracking \cite{luo2021multiple}, etc. 

We first define what we mean by occlusion in the context of optical flow estimation same as GMA \cite{jiang2021learning}, i.e., an occluded point is defined as a 3D point that is imaged in the reference frame but is not visible in the matching frame. 

The optical flow estimation accuracy of non-occluded points is high because non-occluded points exist in the frame pair, and the optical flow value can be directly obtained through local search or global matching. The current average endpoint error of the optical flow of non-occluded points has been reduced to less than 1-pixel \cite{raft}. The optical flow estimation accuracy of the occluded point is much lower. Since the occluded point is not in the next picture, its optical flow can only be reasonably guessed through some indirect means \cite{jiang2021learning}.

To solve the problem of large optical flow prediction error at occluded points, GMA proposed a refinement method for optical flow at occluded points based on the self-similarity assumption, which significantly improves the accuracy of optical flow estimation at occluded points. The self-similarity assumption holds that the motions of a single object (in the foreground or background) are often homogeneous, so the optical flow of the relevant non-occluded points can be used to refine the optical flow of the occluded points on the surface of the same object. The non-occluded points related to the occluded points are called reference point sets. Therefore, the precision of the reference point set is paramount.

GMA and subsequent methods \cite{xu2022GMFlow,sun2022skflow,dong2023rethinking} all find the reference point set through self-attention. However, they cannot avoid drawing attention to the occluded points because relying on the features of a single frame cannot effectively identify and exclude occluded points, making it inevitable that reference points will be contaminated with occluded points, resulting in the inability to provide accurate reference information \textbf{(Untrustworthy)}.
Moreover, due to insufficient constraints, these methods also lead to distracted attention in the trained network, which cannot learn to focus on the most relevant reference point \textbf{(Weakly relevant)}. This results in scattered attention finding reference points that do not provide optimal reference information in common situations such as varying depth of field and non-rigid objects.

Specifically, during the training process, GMA pays attention to 2D motion features, and the penalty is optical flow differences wtih the ground truth (GT), as shown in Figure 2. Since the input of attention is completely different from the GT, it is an indirect constraint. Moreover, there are other complex network layers in between, and the intermediate processes are difficult to explain, resulting in the worst practical performance. Therefore, it is a \underline{\textit{weak, indirect constraint}}. The performance after network training is shown in Figure 1.

GMFlow \cite{xu2022GMFlow} first obtains an optical flow through global matching, $flow_{GM}$, which has small optical flow errors for non-occluded points and large errors for occluded points, as shown in Figure 2.
Then it pays attention to the $flow_{GM}$ and penalizes the difference between the direct result of attention and the optical flow ground truth, which is a direct constraint. This is also an intuitive and easy-to-understand constraint. Because there is a large error in the optical flow of the occluded point in $flow_{GM}$, theoretically it can only learn to focus attention on non-occluded points related to the occluded point to minimize the penalty.

However, during training, the input of the attention matrix provided by GMFlow, $flow_{GM}$, is not completely consistent with the GT, resulting in a \underline{\textit{weak direct constraint}}. There are two reasons for this: First, the optical flow corresponding to non-occluded points in $flow_{GM}$ do not match those in GT, which means that even when the attention is fully focused on the non-occluded point itself, it may not necessarily result in a minimal penalty. This type of training results in a distraction of attention from a non-occluded point that should be completely focused on its own to the area near the point. 
Second, since the optical flow values of the occluded points close to the non-occluded points are close to the trueth value \cite{jing2024yoio}, paying attention to these points will not make the penalty larger, so the attention cannot be effectively restrained from these occluded points. The performance after network training is shown in Figure 1.

Therefore, to completely solve these problems, we integrate the occlusion information into the inference and training process of self-attention, as shown in Figure 2. We employ an online occlusion information recognizer \cite{jing2024yoio}, which does not require occlusion ground truth to supervise the training process.
To enable the network to learn that it should not focus on any occluded points, we designed a repulsion constraint. Based on $flow_{GM}$, we use occlusion information to set the optical flow of occluded points in $flow_{GM}$ to 0 (or other singular values). In this way, during training, the network will receive a large penalty if it focuses on any occluded points, so it is a  \underline{\textit{strong repulsion constraint}} \textbf{(Trustworthy and strongly relevant)}. The performance after network training is shown in Figure 1.

To further enhance the attention ability of the network, we designed attraction constraints based on the principle that it is optimal for the non-occluded point to only focus on itself (especially on the surface of such objects, which have different depths of field or are non-rigid bodies). We use the occlusion information to find the set of non-occluded points so that the attention of these points can be fully focused on themselves, otherwise they will surely be punished. Since the input to the attention matrix and the supervision value are completely consistent, the penalty is minimized only when the non-occluded point focuses on itself, so this is a  \underline{\textit{strong attraction constraint}} \textbf{(The most relevant)}. 
The attention targets and results of the above two constraints belong to the same thing (optical flow), so they are  \underline{\textit{direct strong constraints}}.

We have made the following progress by fully integrating the occlusion information provided by the online occlusion detector into the inference and training process of the network for the first time:

1. Constructing \textbf{occlusion extended feature representations} satisfies the prerequisite that self-attention can identify and remove attention to any occluded points. 

2. Through the strong repulsion constraints designed, the network can learn effective occlusion extended feature representations and only pay attention to strongly relevant non-occlusion points in reasoning. \textbf{(Trustworthy and strongly relevant)}.

3. Through the strong attraction constraint designed, the attention ability training of the network is further strengthened, allowing it to focus only on the most relevant reference points during the reasoning process. \textbf{(The most relevant)}

4. Our method adds \textbf{very few parameters} to the original framework, making it very lightweight. The occlusion information recognizer used is parameter-free, fast in computation, and does not require occlusion labels to supervise network training. These are all very beneficial for reasoning and training networks.

5. Extensive experiments show that our model has \textbf{the greatest cross-dataset generalization}. Our method achieves much greater error reduction, 18.6\%, 16.2\%, and 20.1\% for all points, non-occluded points, and occluded points respectively from the state-of-the-art GMA-base method, MATCHFlow(GMA), on Sintel Albedo pass. Furthermore, our model achieves state-of-the-art performance on the Sintel benchmarks, \textbf{ranking \#1} among all published methods on Sintel clean pass.

\section{Related Work}

\textbf{Flow estimation approach.} The flow estimation approach is fundamental to existing popular optical flow framework \cite{ilg2017flownet,sun2018pwc,hur2019iterative,raft,jiang2021learning,xu2022GMFlow,shi2023flowformer++}, notably the coarse-to-fine method PWC-Net \cite{sun2018pwc} and the iterative refinement method RAFT \cite{raft}. They both perform some sort of multi-stage refinements, either at multiple scales \cite{sun2018pwc} or a single resolution \cite{raft}. For flow prediction at each stage, their pipeline is conceptually similar, i.e., regressing optical flow from a local cost volume with convolutions. To improve the optical flow estimation effect in occlusion areas, GMA proposed a global motion aggregation module based on self-attention embedded in the RAFT architecture. This article makes full use of online occlusion information to further improve performance of self-attention and obtain trustworthy attention.

\textbf{Occlusions information  in optical flow.} Understanding occlusion information is beneficial for improving estimation performance, but obtaining occlusion information during network training and inference is not easy \cite{zhao2020maskflownet,hur2019iterative,jonschkowski2020matters,jeong2022imposing}. On the one hand, it is because obtaining occlusion labels is difficult. On the other hand, it is because the computation of using bidirectional optical flow to estimate occlusion information is large \cite{alvarez2007symmetrical,hur2017mirrorflow,xu2022GMFlow}, which is not suitable for use during network inference. Recently, a method called YOIO has been published, which is an online, parameter-free occlusion estimation method. Recently, YOIO has published an online, parameter-free occlusion estimation method that has low computational complexity, does not require occlusion ground truth for network training, and has a broader range of applications. Therefore, this article relies solely on YOIO for obtaining occlusion information during network training and inference.

\renewcommand\floatpagefraction{.999}
\renewcommand\topfraction{.9}
\renewcommand\bottomfraction{.9}
\renewcommand\textfraction{.1}
\begin{figure*}[htbp]
    \centering
    \includegraphics[width=1.0\textwidth]{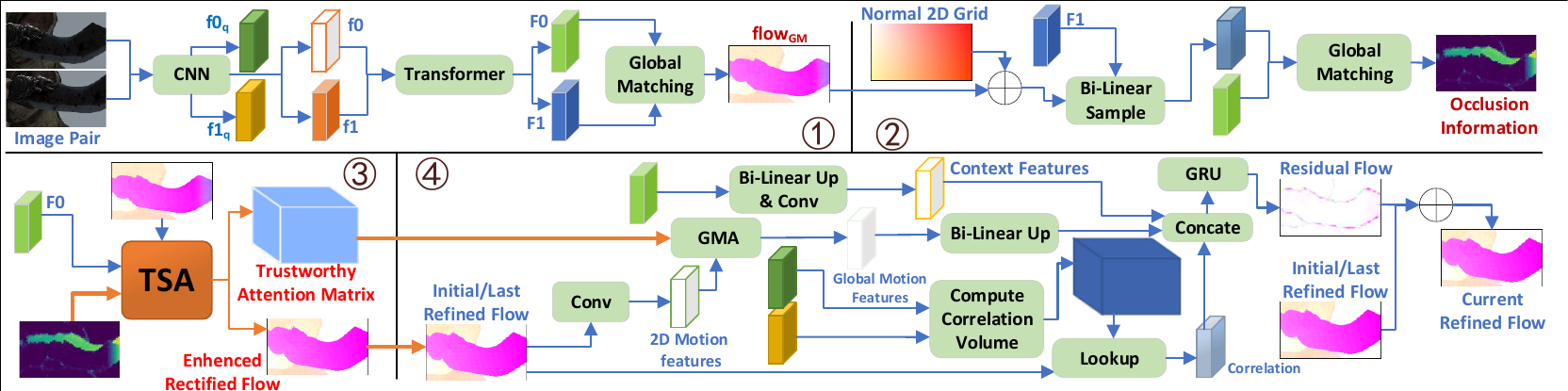}  
    \caption{The overall framework of our method. Our method consists of four stages that are executed sequentially in order. The first stage takes image pairs as input and extracts local image features $f0_q$ and $f1_q$ at 4x downsampling, as well as global image features F0 and F1 at 8x downsampling. It also calculates the $flow_{GM}$. The second stage takes the global image features F0, F1, and $flow_{GM}$ as input to calculate the occlusion information. The third stage uses F0, occlusion information, and $flow_{GM}$ as input to obtain the trustworthy attention matrix and rectified flow. The fourth stage takes F0 upsampled by 2x and convolved as the Context feature, enhanced rectified flow as the initial flow for the Refinement process. It also utilizes $f0_q$ and $f1_q$ to calculate the correlation volume. We replace the attention matrix in GMA with our trustworthy attention matrix.After sufficient iterations of Refinement, the current refined flow is used as the final output.}
    \label{fig3}
\end{figure*}

\section{Method}
\subsection{Overview}

As shown in Figure 3, the reasoning process of our method is divided into four stages. The first stage is feature extraction and initial optical flow acquisition, refer to GMFlow \cite{xu2022GMFlow}. The second stage is to obtain occlusion information, refer to YOIO \cite{jing2024yoio}. The third stage is to perform self-attention that fuses occlusion information to obtain a trustworthy attention matrix and rectified optical flow, refer to Section 3.2. The last stage is to use the above information to refine through loop iteration to obtain the final optical flow, see Section 3.4.

Our method uses occlusion information to add two additional strong direct constraints during the training process, refer to Section 3.3 and 3.4.

\subsection{Occlusion Extended Features}
The occlusion information output from the second stage is denoted as OM. In theory, points with a value of 0 in OM are non-occluded points, and points with a value greater than 0 are occluded points \cite{jing2024yoio}. We concatenate OM after two layers of convolution with visual features of image0 (F0) to obtain the occlusion extended features.

\subsection{Strong Repulsion Constraint}
Since the scale of the feature is only 1/8 of the input image. Therefore, we regard points with values less than 1/8 in OM as non-occluded points, obtaining a non-occluded mask $OM_n$. The non-occluded points in $OM_n$ have a value of 1, and the rest are 0. Using the dot product of $OM_n$ and $flow_{GM}$, we set the optical flow of occluded points to 0, leaving only the optical flow values of non-occluded points, obtaining a Masked $flow_{GM}$, which is used as the input for strong repulsive constraints, as shown in Figure 2. The rest of the steps are the same as GMFlow.

\subsection{Strong Attraction Constraint}
Use $OM_n$ to sample the corresponding sub-attention matrix M from the first two dimensions of the attention matrix for all non-occluded points. We also use $OM_n$ to sample the corresponding sub-normal 2D grid from the normal 2D grid. Then, based on them, we construct strong attraction con-straints, as shown in Figure 2.

\subsection{Iterative Refinement}
We believe that CNNs possess more local detailed infor-mation. Therefore, we utilize CNNs to obtain image features at 4x downsampling for calculating the correlation volume. Correspondingly, we upsample F0 by 2x and convolve it as the context feature. We also upsample the motion features extracted from rectified flow by 2x. We replace the attention matrix in GMA \cite{jiang2021learning} with our attention matrix. The rest is the same as RAFT \cite{raft}.

\subsection{Total Training Loss}
We add the above two strong direct constraints to the original constraints of RAFT to supervise network training. The weight of the new constraints is set to 0.6.

\section{Experiments}
\subsection{Experimental Setup}

\textbf{Datasets and evaluation setup}. Following previous method \cite{raft}, we first train on the Fly-ing-Chairs(Chairs) \cite{dosovitskiy2015flownet} and FlyingThings3D (Things)   datasets \cite{mayer2016large}, and then evaluate Sintel \cite{butler2012naturalistic} and KITTI training sets \cite{menze2015object}. Finally, we perform additional fine-tuning on Sintel training sets and report the performance on the online benchmarks. 

\textbf{Metrics}. We adopt the commonly used metric in optical flow, i.e., the average end-point-error (AEPE), which is the mean $\ell2$ distance between the prediction and ground truth.

\textbf{Implementation details.} We implement our framework in PyTorch \cite{paszke2019pytorch}. The implementation and configuration of our first, third, and fourth stages are consistent with the referenced methods.
We use AdamW \cite{bai2022deep} as the optimizer. We, first train the model on the Chairs dataset for 100K iterations, with a batch size of 9 and a learning rate of 4e-4. We then fine-tune it on the Things dataset for 1500K iterations, with a batch size of 6 and a learning rate of 2e-4. For the final fine-tuning process on Sintel datasets, we further fine-tune our Things model on several mixed datasets that consist of KITTI, HD1K \cite{kondermann2016hci}, FlyingThings3D, and Sintel training sets. We perform fine-tuning for 600K iterations with a batch size of 6 and a learning rate of 2e-4.

\subsection{Comparison of Self-attention Performance}
To ensure the fairness of experiments and test the generalization ability of these methods, we train on both the Chairs and Things datasets and then test on the Sintel dataset.
\renewcommand\floatpagefraction{.9}
\renewcommand\topfraction{.9}
\renewcommand\bottomfraction{.9}
\renewcommand\textfraction{.1}
\setcounter{totalnumber}{50}
\setcounter{topnumber}{50}
\setcounter{bottomnumber}{50}
\begin{figure}[htbp]
    \centering
    \includegraphics[width=8cm,height=9.8cm]{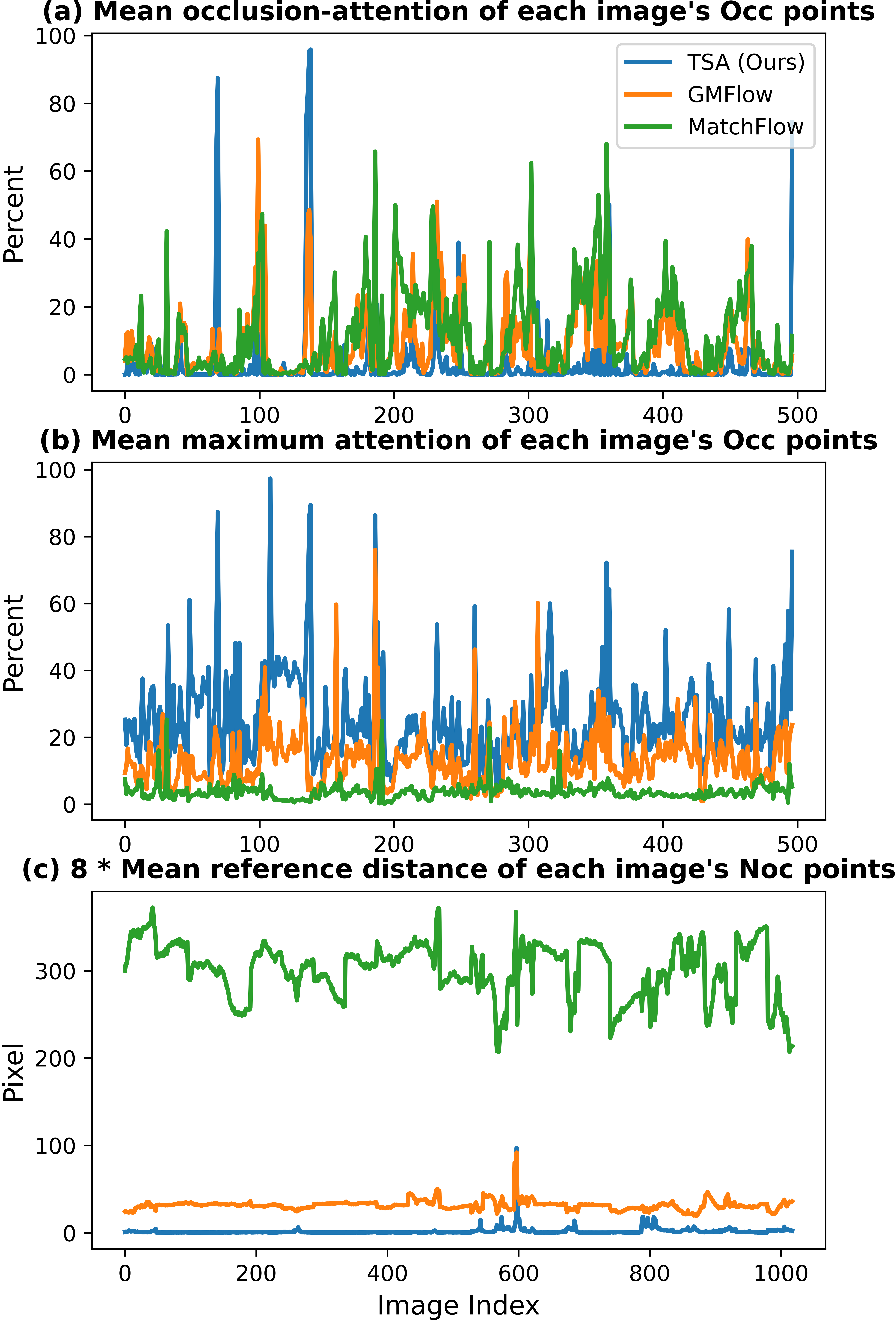}
    \caption{ The Mean of each metric for each possible image on the Sintel Clean Pass. "Noc" means non-occluded. "Occ" means occluded. In (a) (c), the smaller the value, the better. In (b), the larger the value, the better. It can be seen that this method has obvious advantages over the other two methods.}
    \label{fig4}
\end{figure}

We utilize occlussion ground truth (GT) $OM \in R^{H \times W}$ to perform statistical analysis. Since the scale of attention matrix is only 1/8 the size of the input image, we can only downsample the occlussion GTs by 8x. Due to inevitable information distortion after downsampling, we first mark points that are 0 after downsampling as quasi-occluded points. If an occluded (occ) point is surrounded by other quasi-occ points, the quasi-occ point is confirmed as a true occluded point, and all points are marked as $OM_o$. Similarly, we can obtain all true non-occluded points (with a value of 1) by marking them as $OM_n$.

\textbf{MOA.} First, we define the mean occlusion-attention (MOA) to validate the trustworthiness of our attention matrix $M\in R^{(H \times W \times H\times W)}$. The mean occlusion-attention for non-occluded points on any given image can be calculated using formula 1. Replace the "$OM_n$" with "$\neg{OM_{o}}$" in formula 1 to obtain the mean occlusion-attention for occluded points.

\begin{table}
    \centering
    \begin{tabular}{cccc}
        \hline
        MOA(\%) & MATCHFlow(GMA)  & GMFlow & TSA(Ours) \\
        \hline
        Occ    & 6.99     & 4.64     & \textbf{0.08} \\
        \hline
    \end{tabular}
    \caption{The median of the mean occlusion-attention (MOA) of occluded (Occ) points on all images. The best result is styled in bold.}
    \label{tab1}
\end{table}
\begin{table}
    \centering
    \begin{tabular}{cccc}
        \hline
       MRD  & MATCHFlow(GMA)  & GMFlow & TSA(Ours) \\
        \hline
        Noc    & 38.38     & 3.967     & \textbf{0.064} \\
        \hline
    \end{tabular}
    \caption{The median of the mean reference distance (MRD) of non-occluded (Noc) points on all images. The unit of MRD is 1/8 of a pixel. The best result is styled in bold.}
    \label{tab2}
\end{table}
\begin{table}
    \centering
    \begin{tabular}{cccc}
        \hline
      MMA(\%)  & MATCHFlow(GMA)  & GMFlow & TSA(Ours) \\
        \hline
        Noc    & 0.03     & 5.82     & \textbf{97.19} \\
        \hline
    \end{tabular}
    \caption{The median of the mean maximum attention (MMA) of all the non-occluded (Noc) points on all images. The best result is styled in bold.}
    \label{tab3}
\end{table}
\begin{table}
    \centering
    \begin{tabular}{cccc}
        \hline
       MMA(\%) & MATCHFlow(GMA)  & GMFlow & TSA(Ours) \\
        \hline
        Occ    & 3.42     & 12.2     & \textbf{22.5} \\
        \hline
    \end{tabular}
    \caption{ The median of the mean maximum attention (MMA) of occluded (Noc) points on all images. The best result is styled in bold.}
    \label{tab4}
\end{table}

\begin{equation}
    \label{eqa:atten}
    (\sum_{1}^{H \times W}(\sum_{1}^{H \times W}M\cdot \neg{OM_{o}})\cdot OM_{n})/\sum_{1}^{H \times W}OM_{n}
\end{equation}

To make a more fair comparison, we choose the median of the mean of all images as the overall metric. Figure 4 (a) shows the mean of all images, from which it can be seen that there are occasionally large outliers. This is due to image features being poorly extracted in some images, such as those with large areas of textureless regions. To exclude the interference caused by differences in image feature extractors, we choose to compare using the median. More detailed statistics are provided in the appendix.

We have calculated the mean occlusion-attentions for non-occluded points under three methods, which are all less than 1\%. 

The mean occlusion-attention for occluded points is shown in Table 1 and Figure 4(a). It can be seen that the weak direct constraint method performs better than the weak indirect constraint method, but it still cannot avoid focusing on occluded points (\underline{\textit{Untrustworthy}}), while our method barely focuses on Non-occluded points (\textbf{Trustworthy}).

\textbf{MRD and MMA.}In addition, we also analyze the degree of non-occluded points' attention on themselves. We use two metrics: the first is the mean reference distance (MRD), which is the weighted sum of the distances from each point to all other points, with weights M. The second is the mean maximum attention (MMA), which is the mean of the maximum values in the attention maps of all non-occluded points.

Our method has an almost zero mean reference distance and an mean maximum attention close to 100\%, fully demonstrating that our method enables non-occluded points to focus only on themselves (\textbf{The most relevant}), which is significantly different from the other two methods (\underline{\textit{Weakly relevant}}). As shown in Table 2, Table 3 and Figure 4(c).
The mean reference distance metric is not suitable for occluded points because their reference points may include occluded points that are closer to them, and comparing distances is meaningless in this case.

For the mean maximum attention for occluded points, we also have a nearly 200\% improvement (\textbf{Strongly relevant}) compared to GMFlow (\underline{\textit{Weakly relevant}}), as shown in Table 4 and Figure 4(b).

These data, which show \textbf{our TSA $>>$ GMFlow $>$ MATCHFlow(GMA)}, all support our theoretical analysis and prove that our method is more reasonable and effective.

\subsection{Comparison of Optical Flow Estimation Performance}

\begin{table}
    \centering
    \begin{tabular}{cccc}
        \hline
        Type &  {\makecell[c]{GMFlow \\  (AEPE)}}     &  {\makecell[c]{TSA(Ours) \\  (AEPE)}}   & {\makecell[c]{Rel.Impr. \\ (\%)}}       \\
        \hline
        All   & 1.72   & \textbf{1.64}        & 4.25 \\
        Noc   & 0.98   & \textbf{0.95}        & 3.07 \\
        Occ   & 9.71 & \textbf{9.18}           & \textbf{5.49} \\
        \hline
    \end{tabular}
    \caption {APEP of recitified flow on Sintel clean pass, partitioned into occluded (Occ) and non-occluded (Noc) regions. The best results and the largest relative improvement in each dataset are styled in bold.}
    \label{tab5}
\end{table}

\begin{table}
    \setlength{\tabcolsep}{1.3mm}{
    \centering
    \begin{tabular}{ccccc}
        \hline
        Sintel Pass & Type  & {\makecell[c]{MATCHFlow \\ (GMA)(AEPE)}}  &  {\makecell[c]{TSA(Ours) \\  (AEPE)}} & {\makecell[c]{Rel.Impr. \\ (\%)}} \\
        \hline
        \multirow{5}*{Clean}    & Noc     & 0.44     & \textbf{0.39}      & 11.36 \\
        & Occ     & 8.51     & \textbf{7.04}      & 17.27 \\
        & Occ-in  & 6.42     & \textbf{5.68}      & 11.53 \\
        & Occ-out & 9.63     & \textbf{7.28}      & \textbf{24.40} \\
        & All     & 1.03     & \textbf{0.87}      & 15.53 \\
        \hline
        \multirow{5}*{Albedo}    & Noc     & 0.37     &\textbf{0.31}      & 16.22 \\
        & Occ     & 7.98     & \textbf{6.38}      & 20.05 \\
        & Occ-in  & 6.37     & \textbf{5.33}      & 16.33 \\
        & Occ-out & 8.39     & \textbf{6.26}      & \textbf{25.39} \\
        & All     & 0.92     & \textbf{0.75}      & 18.48 \\
        \hline
    \end{tabular}
    \caption{Optical flow error for different Sintel datasets, parti-tioned into occluded (Occ) and non-occluded (Noc) regions. In-frame and out-of-frame occlu-sions are further split and denoted as Occ-in and Occ-out. The best results and the largest relative improvement in each dataset are styled in bold.}
    \label{tab6}
    }
\end{table}

The experimental configuration is the same as 4.2.
We first compared the accuracy of rectified flow. Under the same conditions and use bilinear interpolation to enlarge the rectified flow by 8 times, our method improved accuracy in both occluded ('Occ') and non-occluded ('Noc') regions, with a relatively greater improvement in occluded regions. As shown in Table 5.

In addition, our method also significantly outperforms the latest GMA-based method, MATCHFlow(GMA), in both occluded and non-occluded regions. As shown in Table 6.

Albedo is a series of images that lack any special effects, ensuring that the analysis remains free from interference by exaggerating the differences in performance during image feature extraction, particularly in relation to motion blur. so it is most suitable for comparison here. Our method achieves much greater error reduction, 18.6\%, 16.2\%, and 20.1\% for all points, non-occluded points, and occluded points respectively.

On the clean pass, the improvement margin is smaller (although still significant), which we attribute to the advantage of MATCHFlow(GMA) in feature extraction. Since our model has a total parameter count of less than 800M, while  MATCHFlow(GMA) has a total parameter count of 15.4M, it should have a stronger ability in image feature extraction.

Both sets of experiments prove that our method has significantly better performance in optical flow estimation, which once again proves that our method is more reasonable and effective.

\subsection{Qualitative Results}

\begin{table}
    \centering
    \begin{tabular}{cccc}
        \hline
        \multirow{2}{*}{Training Data} & \multirow{2}{*}{Method} & \multicolumn{2}{l}{Sintel(train)} \\ \cline{3-4} 
                               &                         & Clean           & Final           \\
        \hline
        \multirow{8}*{C+T} & RAFT     & 1.43     & 2.71 \\
        & GMA      & 1.3      & 2.74 \\
        & GMFlow   & 1.08     & 2.48 \\
        & SKFlow   & 1.22     & 2.46 \\
        & MATCHFlow(RAFT)   & 1.14     & 2.61 \\
        & MATCHFlow(GMA)   & 1.03     & 2.45 \\
        & FlowFormer++  & 0.9      & \textbf{2.3} \\
        & TSA(Ours)      & \textbf{0.87}     & 3.01 \\
        \hline
        \multirow{8}*{C+T+S}   & RAFT     & 1.61*     & 2.86* \\
        & GMA      & 1.39*      & 2.17* \\
        & GMFlow   & 1.74     & 2.9 \\
        & SKFlow   & 1.28*     & 2.23* \\
        & MATCHFlow(RAFT)   & 1.33*     & 2.64* \\
        & MATCHFlow(GMA)   & 1.15     & 2.77 \\
        & FlowFormer++  & 1.07      & \textbf{1.94} \\
        & TSA(Ours)      & \textbf{0.99}     & 2.29 \\
        \hline
    \end{tabular}
    \caption{Quantitative results on Sintel datasets. "C + T" refers to results that are pre-trained on the Chairs and Things datasets. "S + K + H" refers to methods that are fine-tuned on the Sintel and KITTI datasets, with some also fine-tuned on the HD1K dataset. Parentheses denote training set results and bold font denotes the best result. * indicates evaluating with RAFT's "warm-start" strategy.}
    \label{tab7}
\end{table}
We have compared our method with the state-of-the-art methods including RAFT, GMA, GMFlow, MATCHFlow, SKFlow \cite{sun2022skflow}, and FlowFormer++ \cite{shi2023flowformer++}.

As shown in Table 7, we have the greatest performance on Sintel clean pass. It is significantly improved compared to MATCHFlow. It proves that our method is much more effective than other methods.

It is worth noting that in the final pass, our method performed significantly worse than MATCHFlow on the "C+T" experiment. This is because our method relies on the performance of the occlusion recognizer. As shown in Figure 5, due to the addition of difficult special effects such as motion blur in the Final pass image, the occlusion recognizer failed to successfully identify the occlusion information. However, after training on "C+T+S+K+H", the accuracy of occlusion recognition increased, and the error of optical flow estimation also decreased significantly. This is because transformer-based occlusion recognizers require large amounts of data training to improve performance \cite{xu2022GMFlow}. Therefore, our method later surpassed MATCHFlow. FlowFormer++ has 18.2M parameters, and we believe that it still has the advantage of having more parameters. The reason is the same as subsection 4.3.

\subsection{Ablation Study}

\renewcommand\floatpagefraction{.9}
\renewcommand\topfraction{.9}
\renewcommand\bottomfraction{.9}
\renewcommand\textfraction{.1}
\setcounter{totalnumber}{50}
\setcounter{topnumber}{50}
\setcounter{bottomnumber}{50}
\begin{figure*}[htbp]
    \centering
    \includegraphics[width=17cm,height=3cm]{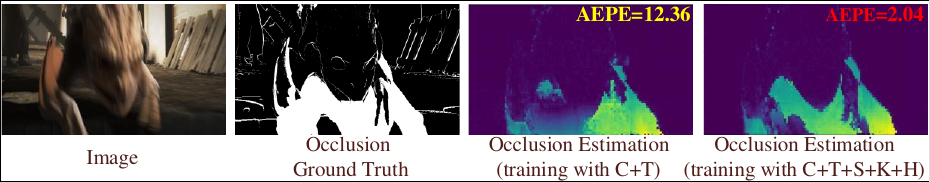}
    \caption{The performance of our method depends on the performance of the occlusion information detector.When confronted with input images suffering from suboptimal qualities such as significant motion blur and a lack of comprehensive training data, the detector encounters challenges in accurate occlusion recognition, leading to substantial optical flow prediction errors. However, by increasing the amount of training data, the occlusion recognition effectiveness improves, resulting in smaller optical flow prediction errors.}
    \label{fig5}
\end{figure*}

\begin{table}
    \centering
    \begin{tabular}{ccc}
        \hline
        Approach & AEPE  &  {\makecell[c]{Rel.Impr. \\ (\%)}} \\
        \hline
        GMFlow + GMA    & 1.95     & /  \\
        + Occluded extended Feature maps    & 1.89     & 3.08 \\
        ++ Strong Repulsion loss    & 1.83      & 3.17  \\
        +++ Strong Attraction loss    & 1.8      & 1.64  \\
        \hline
    \end{tabular}
    \caption{ Ablation study. It shows that as each component is incrementally added, the overall performance gradually improves. Ultimately, the cumulative improvement amounts to a significant 7.8\%, thereby validating the efficacy and essential nature of each individual component. One or more “+” indicate stacking on the previous methods.}
    \label{tab8}
\end{table}

To save time, we only trained on the Chairs dataset and tested on the Sintel dataset.

First, we use "GMFlow + GMA" as the baseline. Then, we add 0ccluded extended features, followed by the strong repulsion constraint and finally the Strong Attraction constraint. It can be seen that the performance gradually improves, with an ultimate improvement of 7.8\%. 
To analyze the effect of adding various components more easily, we drew the mean reference distance after fine-tuning on the Sintel dataset, as shown in Figure 6. We can see that after adding the strong repulsion constraint, the non-occluded points have basically learned to attend to themselves, indicating that the network has learned a good occlusion-aware feature representation. After adding the strong attraction constraint, visually, almost all non-occluded points have learned to reference themselves only.

Experiments have proved that our method is more reasonable, effective, and easy to understand. 

\renewcommand\floatpagefraction{.9}
\renewcommand\topfraction{.9}
\renewcommand\bottomfraction{.9}
\renewcommand\textfraction{.1}
\setcounter{totalnumber}{50}
\setcounter{topnumber}{50}
\setcounter{bottomnumber}{50}
\begin{figure}[htbp]
    \centering
    \includegraphics[width=8.5cm,height=4.5cm]{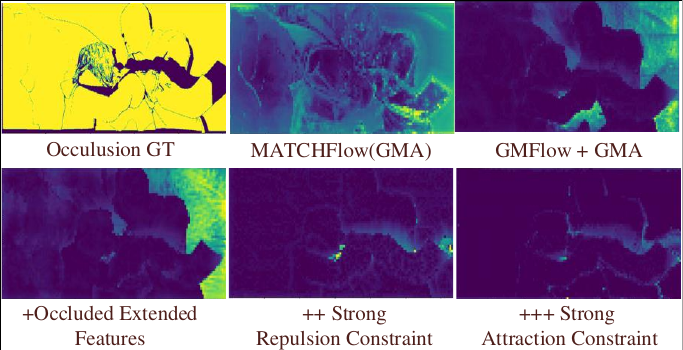}
    \caption{The mean reference distance of all points on the image. It shows that after adding the Strong Repulsion constraint, the non-occluded points have basically learned to attend to themselves only, indicating that the network has learned a good occlusion-aware feature representation. After adding the Strong Attraction constraint, visually, almost all non-occluded points have learned to reference themselves only. One or more “+” indicate stacking on the previous method.}
    \label{fig6}
\end{figure}

\section{Conclusion}

We are the first to use online occlusion recognition to identify occlusion information and integrate this information into the training and inference process of self-attention. This allows the network to learn to focus only on the most relevant information (Trustworthy). Our method adds very few network parameters to the original framework, making it very lightweight. Extensive experiments show that our model has the greatest cross-dataset generalization. Our method achieves much greater error reduction, 18.6\%, 16.2\% and 20.1\% for all points, non-occluded points, and occluded points respectively from the state-of-the-art GMA-base method, MATCHFlow(GMA), on Sintel Albedo pass. Furthermore, our model achieves state-of-the-art performance on the Sintel benchmarks, ranking \#1 among all published methods on Sintel clean pass.
\bibliographystyle{named}
\bibliography{ref}

\end{document}